\documentclass{article}

\usepackage{arxiv}
\usepackage{tikz}
\usetikzlibrary{calc, backgrounds, positioning}

\usepackage[utf8]{inputenc} % allow utf-8 input
\usepackage[T1]{fontenc}    % use 8-bit T1 fonts
\usepackage{hyperref}       % hyperlinks
\usepackage{url}            % simple URL typesetting
\usepackage{booktabs}       % professional-quality tables
\usepackage{amsfonts}       % blackboard math symbols
\usepackage{nicefrac}       % compact symbols for 1/2, etc.
\usepackage{microtype}      % microtypography
\usepackage{lipsum}
\usepackage{graphicx}
\graphicspath{ {./images/} }

\title{Automated Fact-Checking: A Survey}

\author{\textbf{Xia Zeng}\\
Queen Mary University of London, UK
\and \textbf{Amani S. Abumansour}\\
Queen Mary University of London, UK\\
Taif University, Saudi Arabia
\and \textbf{Arkaitz Zubiaga}\\
Queen Mary University of London, UK}

\begin{document}
\maketitle

\begin{abstract}
As online false information continues to grow, automated fact-checking has gained an increasing amount of attention in recent years. Researchers in the field of Natural Language Processing (NLP) have contributed to the task by building fact-checking datasets, devising automated fact-checking pipelines and proposing NLP methods to further research in the development of different components. This paper reviews relevant research on automated fact-checking covering both the claim detection and claim validation components.

\end{abstract}


\section{Introduction} \label{introduction}
While online content continues to grow unprecedentedly, the spread of false information online increases the potential of misleading people and causing harm. This leads to an increasing demand on fact-checking, i.e. a task consisting in assessing the truthfulness of a claim \cite{vlachos_fact_2014}, where a claim is defined as `a factual statement that is under investigation' \cite{hanselowski_machine-learning-based_2020}. A number of fact-checking organisations have been founded in recent years, e.g. FactCheck, PolitiFact, Full Fact, Snopes, Poynter and NewsGuard. Fact-checkers have conducted laborious manual fact-checking, i.e. familiarise with the topic, identify the claim, aggregate evidence, check source credibility, verify the claim and its reasoning chain and check for fallacies \cite{hanselowski_machine-learning-based_2020}. However, the speed and efficiency of manual fact-checking cannot keep up with the pace at which online information is posted and circulated. The journalism community can benefit from tools that, at least partially, automate the fact-checking process \cite{cohen_computational_2011, claimbuster2017, thorne_automated_2018, konstantinovskiy_towards_2020}, particularly by automating more mechanical tasks, so that human effort can instead be dedicated to more labour-intensive tasks \cite{babakar2016state}. Restricting claims to those that are objectively fact-checkable makes the automation task more realistically achievable while reducing the volume of content needing manual fact-checking. Furthermore, recent progress in the fields of natural language processing (NLP), information retrieval (IR) and big data mining has demonstrated the potential for efficiently processing large-scale textual information online, which inspires automated fact-checking.

Researchers have developed valuable fact-checking datasets, pipelines and models, an effort which has also been supported by shared tasks, including RumourEval \cite{derczynski_semeval-2017_2017, gorrell_rumoureval_2018}, CLEF CheckThat! \cite{nakov_overview_2018, elsayed_overview_2019, barron-cedeno_overview_2020}, ClaimBuster \cite{hassan_claimbuster_2017}, FEVER \cite{thorne_fever_2018, thorne_fever20_2019}, SCIVER \cite{wadden_fact_2020}, Fake News Challenge \cite{pomerleau2017fake}, and HeroX fact checking challenge \cite{francis2016herox}. 

With different major concerns, proposed pipelines take various forms. For instance, ClaimBuster \cite{hassan_claimbuster_2017} designed a comprehensive pipeline of four components to verify web documents: a claim monitor that performs document retrieval; a claim spotter that performs claim detection; a claim matcher that matches a detected claim to fact-checked claims; a claim checker that performs evidence extraction and claim validation. A similar pipeline was proposed by CLEF CheckThat! \cite{nakov2021clef}, which in its 2021 edition included three subtasks: first, perform claim detection to detect claims that are check-worthy; second, determine whether a claim has been previously fact-checked; and third, perform claim validation to determine the factuality of the detected claims. While some pipelines include claim detection, some are only designed to tackle claim validation, e.g. FEVER \cite{thorne_fever_2018, thorne_fever20_2019} and SCIVER \cite{wadden_fact_2020},\footnote{The task of claim validation is referred to as fact-checking by some papers in the literature.} assuming check-worthy claims are already at hand. Figure \ref{fig:pipeline} depicts a comprehensive fact-checking pipeline as discussed in this survey and consisting of two components: (1) a claim detection component, which looks for claims that need checking and tries to find matches between claims when they are related to the same fact-check, and (2) a claim validation component, which retrieves the documents and rationales that can serve as evidence to fact-check a claim and ultimately performs the verification task, producing a verdict.

\usetikzlibrary{calc, backgrounds, positioning}

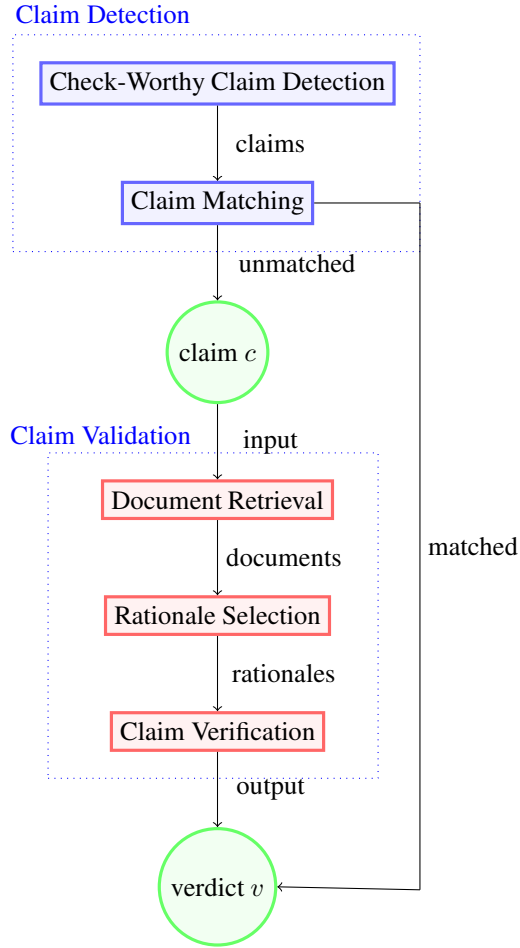
\begin{figure}
\centering
\begin{tikzpicture}[
roundnode/.style={circle, draw=green!60, fill=green!5, very thick, minimum size=7mm},
squarednodeb/.style={rectangle, draw=blue!60, fill=blue!5, very thick, minimum size=5mm},
squarednode/.style={rectangle, draw=red!60, fill=red!5, very thick, minimum size=5mm},
]
%Nodes
\node[squarednodeb]      (CD)       {Check-Worthy Claim Detection};
\node[squarednodeb]      (CM)       [below=of CD]{Claim Matching};
\node[roundnode]        (claim)       [below=of CM]{claim $c$};
\node[squarednode]      (DR)    [below=of claim] {Document Retrieval};
\node[squarednode]      (RS)       [below=of DR] {Rationale Selection};
\node[squarednode]      (CV)       [below=of RS] {Claim Verification};
\node[roundnode]        (verdict)       [below=of CV] {verdict $v$};

%Lines
\draw (CD.south) 
      edge[->] node[xshift=2em]{claims} 
      (CM.north);
\draw (CM.east) -- ($(CM.east) + (4em, 0em)$);
\draw ($(CM.east) + (4em, 0em)$) 
      edge[-] node[xshift=2em]{matched} 
      ($(CM.east) + (4em, -26em)$);
\draw ($(CM.east) + (4em, -26em)$) 
      edge[->]
      (verdict.east);
\draw (CM.south) 
      edge[->] node[xshift=3em]{unmatched} 
      (claim.north);
\draw (claim.south) 
      edge[->] node[xshift=2em]{input} 
      (DR.north);
\draw (DR.south) 
      edge[->] node[xshift=2.5em]{documents} 
      (RS.north);
\draw (RS.south) 
      edge[->] node[xshift=2.5em]{rationales} 
      (CV.north);
\draw (CV.south) 
      edge[->] node[xshift=2em]{output} 
      (verdict.north);

% draw dotted lines
\begin{pgfonlayer}{background}
\draw [join=round,blue,dotted] ($(CD.north west) + (-1em, 1em)$) rectangle ($(CM.south east) + (4em, -1em)$) node[xshift=-12em, yshift=9em]{Claim Detection};
\draw [join=round,blue,dotted] ($(DR.north west) + (-2em, 1em)$) rectangle ($(CV.south east) + (2em, -1em)$) node[xshift=-10.5em, yshift=13em]{Claim Validation};
\end{pgfonlayer}
\end{tikzpicture}
\caption{An Overview of Automated Fact-checking System.}
\label{fig:pipeline}
\end{figure}

In this paper, we present a survey on automated fact-checking with special focus on claim detection and claim validation, and is structured as follows. Section \ref{claim detection} focuses on the task of detecting check-worthy claims, which is the very first task of a comprehensive automated fact-checking pipeline. It also contains a brief overview of claim matching. Section \ref{claim validation} presents the task of claim validation, which typically involves addressing evidence retrieval and claim verification together. In section \ref{discussion}, we discuss advantages and drawbacks of current automated fact-checking pipelines, with a focus on current challenges and promising future directions. Section \ref{related} presents closely related NLP tasks. Conclusions are drawn in Section \ref{conclusions}.

\section{Claim Detection} \label{claim detection}
Claim detection plays a crucial role in automated fact-checking systems as all other components need to rely on the output of this stage. It aims to relief the burden of identifying claims for fact-checkers and help them by minimising the volume of online content they need to deal with. 

\subsection{Approaches}

The claim detection component is responsible for selecting claims that need to go through the rest of the fact-checking pipeline due to needing to be checked, i.e. needing verification. For instance, a factual statement such as ``He voted against the first gulf war'' can be deemed a claim that should be fact-checked. In contrast, a piece of opinion such as ``I think it's time to talk about the future'' is not a claim that should be fact-checked \cite{claimbuster2017}.

Going further, one can also distinguish between check-worthy vs non-check-worthy claims \cite{nakov2021clef}, as cases that being claims are worthy or not fact-checking. For example, one could argue that ``the government invested more than 10 billion last year in education'' is a claim that is worthy of fact-checking, whereas a claim such as ``my friend had a coffee this morning for breakfast'' may not be worthy of fact-checking.

Researchers typically formulate the problem as one having a set of sentences as input (e.g. originating from a debate or conversation), and is tackled as a classification task, where a binary decision is made on whether each input sentence constitutes a claim or not, or a ranking task, where input sentences are ranked by check-worthiness, hence prioritising top claims on top positions of the list. 

\subsection{Datasets}
In recent studies, several datasets were built with the purpose of enabling training machine learning models to predict check-worthy claims, as shown in Table \ref{tab:CDdataset}. The vast majority of datasets cover sentences pertaining to the political domain, as a result of events that synchronously occur with the US elections. In contrast, the CheckThat! Lab released English and Arabic datasets that contain a small number of instances related to COVID-19 in early 2020.

\begin{table}[h!]
 \caption{Check-worthiness claim detection Datasets}
  \centering
  \begin{tabular}{p{5.5cm}p{3cm}p{4cm}p{2cm}}
    \toprule
    Name     & Size     & Annotation type    & Language \\
    \midrule
ClaimBuster \cite{claimbuster2020}
    &23,533 Sentences 
        & Manual
            & English\\
\addlinespace
CW-USPD-2016 \cite{gencheva2017context}
    &5,415 Sentences 
        & From Existing Annotation
            & English\\
\addlinespace
TATHYA
\cite{tathya}
    &15,735 Sentences
        &From Existing Annotation
            &English\\
\addlinespace
Konstantinovskiy et al.,
\cite{konstantinovskiy_towards_2020}
    &5,571 Sentences
        &Based on the annotation scheme, sentences labelled into 7 categories then grouped into 2 categories 
            & English\\
\addlinespace
CT-CWC-18 \cite{ct18t1}
    &8,946 (En),7,254 (Ar) sentences
        &From Existing Annotation
            &English/Partially translated to Arabic\\
\addlinespace
CT19-T1 \cite{ct19t1}
    &23,500 sentences
        &From Existing Annotation
            &English\\
Shaar et al., \cite{ct20en}
    &962 tweets
        &Manual
            &English\\
\addlinespace
CT20-AR
\cite{ct20ar}
    &7.5K tweets
        &Manual
            &Arabic\\

\addlinespace
TrClaim-19 
\cite{trclaim}
    &2287 tweets
        &Manual
            &Turkish\\
\addlinespace
FactRank 
\cite{factrank}
    &7037 sentences
        &Manual
            &Dutch\\
    \bottomrule
  \end{tabular}
  \label{tab:CDdataset}
\end{table}

In addition, publicly available datasets have a variety of sizes. For instance, ClaimBuster \cite{claimbuster2020} and CT19-T1 \cite{ct19t1} are the largest datasets, while CW-USPD-2016 \cite{gencheva2017context}, CT-CWC-18 \cite{ct18t1}, CT20-AR \cite{ct20ar}, and FactRank \cite{factrank} are a degree of magnitude smaller, followed by other smaller datasets.

Datasets have binary-class for either single-label or multi-labels, depending on the annotation process. The annotation process comes in diverse types. Some datasets are automatically built by collecting claims from fact-checking websites, while other datasets rely on manual annotations given specific definitions of check-worthiness. Crowd-sourcing platform has also been demonstrated to be helpful \cite{claimbuster2015}.

Moreover, the majority of datasets are available in the English language as opposed to a smaller number of datasets in the Arabic language. Most of these Arabic datasets are generated from translations of originally English datasets, except for CT20-AR, which is originally Arabic content. In addition, there is one dataset in Dutch and another one in Turkish, while datasets in other languages are not yet available.

\subsection{Check-Worthy Claim Detection}
ClaimBuster is the first automated fact-checking system that consists of integrated components tackling the entire fact-checking pipeline, starting off from the claim detection component. Its claim detection component called ``claim spotter'' classifies input sentences into one of (1) a factual claim, (2) an unimportant factual claim, or (3) a non-factual claim. This in turn assists fact checkers by prioritising the most check-worthy claims by ranking them based on accuracy measures such as Precision at k (P@K) \cite{hassan2017claimbuster}. To develop this, a multi-class Support Vector Machine (SVM) classifier was built which used features such as bags-of-words, Part-Of-Speech (POS) tags, and Entity Types (ET). The model achieved competitive performance and was considered as the baseline to beat in subsequent works \cite{claimbuster2017}.

Another model called ``CNC'' (i.e. ``Claim/not Claim'') \cite{konstantinovskiy_towards_2020} builds on top of InferSent embeddings \cite{conneau2017supervised}, combining them with part-of-speech tags and named entities found in texts, which are fed to a Logistic Regression classifier. Authors of CNC had as their main goal the improvement of the recall score achieved by their system, arguing that fact-checkers don't want to miss out any claims (no false negatives) while they can deal with some false positives. While improving in terms of recall, CNC also achieved superior performance in F1 score.

Apart from traditional machine learning models, neural networks have also been studied for the claim detection task. For example, in the CheckThat! Lab 2019 shared task, LSTM neural networks and Feed forward neural networks were the most effective models used by the top two participants. Along with the use of neural networks, top participants also showed the usefulness of context (i.e. surrounding sentences) in improving claim detection performance \cite{elsayed_overview_2019}. The use of context was studied in more detail in another work conducted outside the shared task, in this case by \cite{cd}. They studied the inclusion of context and discourse features along with sentence-level features. They used a Feed-Forward Neural Network (FNN) as the model, which was then evaluated as a ranking task, proving the effectiveness of context and discourse features.

While all aforementioned works focused on English claims, there have also been efforts in other languages. For example, the ClaimRank model \cite{claimrank} was tested on Arabic claims (translated from claims originally in English). The Arabic claim detection model used Farasa \cite{abdelali2016farasa} for tokenization, part-of-speech (POS) tagging, as well as MUSE embeddings. The first experiments on original Arabic data (rather than translated) were conducted in the CheckThat! 2020 shared task. Most participants proposed approaches fine-tuning pre-trained language models. For instance, the top-performing participant fine tuned AraBERT v0.1 with neural networks \cite{accenture20}. Likewise, \cite{bigir20} fine-tuned multilingual BERT (mBERT) with different classification models. Another recent effort, called FactRank \cite{factrank}, focused on claim check-worthiness detection for the Dutch language, in this case using a convolutional neural network (CNN) along with Platt scaling for an SVM model and a softmax to obtain the degree of check-worthiness.

Given that methods for claim detection have been applied in different settings or on different datasets, it is difficult to establish what the state-of-the-art model is today. However, the best way of determining the best-performing model is to look at the leaderboard in the CheckThat! shared task of the most recent edition.

\subsection{Claim Matching}

Another task that has recently emerged is claim matching, also referred to as identifying previously fact-checked claims. For a claim spotted in the claim detection component, claim matching consists in determining whether this is a claim that exists in the database and can be resolved by a previous fact-check. The task is formulated as follows: given a check-worthy claim as input, and having a database of previously fact-checked claims, it consists in determining if any of the claims in the database is related to the input; in this case, the new claim would not need fact-checking again, as it was fact-checked in the past. It is normally framed as a ranking task, where claims in the database are ranked based on their similarity to the input claim \cite{shaar_that_2020}. This task comes right after the claim detectino component, to determine if the claim is new, and can help avoid the need for running the claim validation component for a particular claim when it is found in the database.

There are two released datasets: one based on PolitiFact and the other based on Snopes. Initial explorations were conducted on using BM25 \cite{robertson_okapi_1994} and BERT-based models respectively as well as building a SVM reranker with features from both approaches \cite{shaar_that_2020}. Otherwise, CLEF2020-CheckThat! held a shared task on Verified Claim Retrieval which uses the Snopes dataset. While the baseline system is a simple BM25 system, shared task participants explored various scoring functions, including unsupervised approaches such as Terrier and Elastic Search scores, classic supervised models such as SVM and various BERT-based models \cite{shaar_overview_nodate}. Buster.ai, the winning team, fine-tune a RoBERTa \cite{liu_roberta_2019} model on the task which was first fine-tuned on other fact-checking datasets \cite{bouziane_team_nodate}. Team UNIPI-NLE, achieving close performance to the winning team, performed two cascade fine-tunings on a sentenceBERT \cite{reimers_sentence-bert_2019} model \cite{passaro_unipi-nle_nodate}.

\section{Claim Validation} \label{claim validation}

As a component of the automated fact-checking pipeline, claim validation is formulated as `the assignment of a truth value to a claim made in a particular context' \cite{vlachos_fact_2014}. 

\subsection{Approaches}
In order to fulfill the task of claim validation, two different major approaches to verification have emerged: 1) the claim is verified against textual references such as documents from Wikipedia \cite{thorne_fever_2018, thorne_fever20_2019}; 2) the claim is verified against existing knowledge bases \cite{shi_discriminative_2016, syed_unsupervised_2019}. Both approaches assume their references are reliable. The first approach may limit evidence to only trusted resource such as Wikipedia, fact-checking websites, peer-reviewed academic papers, and government documents, achieving substantial coverage of information. However, the second approach faces bigger challenges in terms of coverage of reliable information. Existing knowledge bases tend to be too small to cover sufficient information for claim validation purposes \cite{mendes_dbpedia_2012, azmy_farewell_2018, pellissier_tanon_yago_2020}. Attempts have been made to automatically populate knowledge bases \cite{nakashole_real-time_2012, adel_deep_2018, balog_populating_2018, mesquita_knowledgenet_2019} but this method has the risk of further introducing unreliable noise and makes it harder to maintain the knowledge bases. Due to its maturity and reliability, our survey focuses on the first approach.

There have been a number of shared tasks focused on claim validation in slightly different ways. One of the major differences is whether the final verification step is reliant on previously identified pieces of evidence (such as Wikipedia documents or scientific articles) or it is instead reliant on the stances expressed by users (for example by aggregating supporting or opposing stances towards a story in social media). Of those relying on evidence, well-known shared tasks include FEVER \cite{thorne_fever_2018} and SCIVER \cite{wadden_fact_2020}, both of which perform different forms of evidence retrieval first and then perform claim validation based on that evidence. On the other hand, both UKP Snopes \cite{hanselowski_richly_2019} and RumourEval \cite{derczynski2017semeval,gorrell_rumoureval_2018} proposed to tackle the task by retrieving texts relevant to a story, determining the stance of those texts afterwards, to ultimately classify the veracity value of the story.

\subsection{Datasets}

The NLP community has developed valuable datasets to progress research in automated claim validation, though with common issues of being synthetic and imbalanced. As shown in Table \ref{tab:CVdataset}, recent datasets are not only growing in size, but they also attempt to capture naturally occurring sentences, include context and metadata, cover different domains, and offer evidence chains. 

\begin{table}[h!]
 \caption{Claim Validation Datasets}
  \centering
  \begin{tabular}{p{6.3cm}p{2.2cm}p{2.5cm}p{6cm}}
    \toprule
    Name     & \# of Claims or Claim-Evidence Pairs     & Domains    & Details \\
    \midrule
    PolitiFact \cite{vlachos_fact_2014} & 106 claims & Politics & Very small; metadata and evidence of various forms \\
    Emergent \cite{ferreira_emergent_2016} & 300 claims & News & Very small; 2595 associated documents \\
    LIAR \cite{wang_liar_2017} & 12,836 claims & Politics & Medium; metadata \\
    Snopes \cite{popat_where_2017} & 4,956 claims & Snopes website & Medium; 30 Google retrieved documents for each claim \\
    FEVER \cite{thorne_fever_2018} & 185,445 claims & Wikipedia & Big; associated Wikipeida evidence \\
    LIAR-PLUS \cite{alhindi_where_2018} & 12,836 claims & Politics & Medium; automatically extracted justifications \\
    Perspectrum \cite{chen_seeing_2019} & 907 claims & Debates & Small; evidence and perspectives \\
    UKP Snopes \cite{hanselowski_richly_2019} & 6,422 claims & Snopes website & Medium; associated evidence \\
    MultiFC \cite{augenstein_multifc_2019} & 34,918 claims & Fact-checking websites & Medium; metadata and 10 Google retrieved webpages for each claim \\
    Scifact \cite{wadden_fact_2020} & 1,409 claims & Scientific papers & Small; associated documents \\
    PolitiHop \cite{ostrowski_multi-hop_2020} & 500 claims & Politics & Very small; evidence chains for multi--hop reasoning \\
    WikiFactCheck-English \cite{sathe_automated_2020} & 124,821 claims & Wikipedia & Big; context and evidence \\
    Climate-FEVER \cite{diggelmann_climate-fever_2021} & 1,535 claims & Climate & Medium; 7,675 claim-evidence pairs with climate related claims verified against Wikipedia evidence\\
    COVID-Fact \cite{saakyan_covid-fact_2021} & 4,086 claims & COVID-19 & Medium; 1,296 supported claims from r/COVID19 subreddit and 2,790 automatically generated refuted claims \\
    Vitamin-C \cite{schuster_get_2021} & 488,904 pairs & Wikipedia & Big; contrastive evidence from Wikipedia edits \\
    FEVEROUS \cite{aly_feverous_2021} & 87,026 claims & Wikipedia & Biggest; evidence collected from both structured and unstructured information on whole Wikipedia \\
    \bottomrule
  \end{tabular}
  \label{tab:CVdataset}
\end{table}

\subsection{Evidence Retrieval}

Evidence retrieval is conventionally addressed in two steps: document retrieval and rationale selection. Document retrieval is the task of retrieving relevant documents that supports the prediction of a claim's veracity. Rationale selection is the task of selecting directly relevant sentences out of the retrieved documents to get final supporting evidence for claim verification.

\paragraph{Document Retrieval}

Deeply influenced by information retrieval research, the majority of work in the literature addresses it as a ranking problem consisting in retrieving the top k documents. Various combination of Named Entities, Noun Phrases and Capitalised Expressions from the claim were used to query search APIs such as Google or Wikipedia and search servers \cite{thorne_fact_2018}, when participating in the FEVER shared task. Metadata such as page viewership statistics is helpful to rank webpages \cite{nie_combining_2018}. However, when search engines are not available, such as in the SCIVER shared task, the majority of effort goes into exploring similarity metrics that are used as a proxy to determine the documents' relevance to a claim. TF-IDF similarity is a common baseline \cite{wadden_fact_2020, malon_team_2018} and BM25 \cite{robertson_okapi_1994} is demonstrated to be effective \cite{pradeep_scientific_2020}. When dealing with a specific domain, in-domain word embeddings are also a promising option, e.g. BioSentVec \cite{chen_biosentvec_2019} for the SCIFACT dataset \cite{li_paragraph-level_2021}.

Instead of completely relying on unsupervised methods, improvements have been achieved by reranking based on supervised learning on top of a large number of retrieved documents \cite{pradeep_scientific_2020}.

\paragraph{Rationale Selection}
Keyword matching, sentence similarity scoring and supervised ranking are common approaches to rationale selection \cite{thorne_fact_2018}. Similar to document retrieval, attempts typically use one of these approaches or a combination of them to get a ranking score and select top $k$ sentences as rationale with a manual choice of the $k$ value \cite{pradeep_scientific_2020}. 

Most of studies in the literature conduct evidence retrieval by addressing document retrieval and rationale selection in a pipeline manner, which ignores valuable information across sentences.

\subsection{Claim Verification}
Claim verification is commonly addressed as a text classification task by NLP researchers. Given a claim under investigation and its retrieved evidence, models need to reach a verdict of the claim, which may be `SUPPORT', `CONTRADICTION' or `NOT ENOUGH INFORMATION'. Some other datasets \cite{hanselowski_richly_2019, wang_liar_2017} include other labels such as `mostly-true', `half-true', `pants-fire', `most false', `most true' and `other', whose finer granularity is more difficult to tackle through automated means and are sometimes collapsed into fewer labels. An important observation here is the difference in the types of labels used by different studies. Some studies rely on truth values (e.g. true, false, half-true), determining the veracity value of a claim. Others refer to the concept of support instead (i.e. support, contradict), which instead determine whether there is an agreement between the claim and the reference. The latter avoids making an explicit connection with truthfulness, looking instead at the alignment of a claim with respect to a given reference.

The task of claim verification may be essentially addressed as a Recognising Textual Entailment (RTE) task, i.e. `deciding, given two text fragments, whether the meaning of one text is entailed (can be inferred) from another text' \cite{dagan_recognizing_2009} or a Natural Language Inference (NLI) task, i.e. `characterising and using semantic concepts of entailment and contradiction in computational systems' \cite{bowman_large_2015}. 

Given that claim verification is predominantly addressed as a RTE or NLI task, we present a brief overview of them below. The RTE task, which dates back to 2005, focuses on detecting whether the hypothesis $h$ is entailed by a given text $t$ or not, which corresponds to `SUPPORT' or not. Proposed models may take a linguistic approach, a statistical approach, a machine learning approach or a hybrid version of them. The NLI task, equipped with many large-scale labelled datasets, has powered large neural models to be the dominant approach. State-of-the-art models are large pre-trained language models that are fine-tuned on large NLI datasets. 

\paragraph{Recognising Textual Entailment (RTE)}
Textual entailment is defined as a relation between a text $T$ and a hypothesis $H$. Formal semantics defines that a text $t$ entails hypothesis $h$ if $h$ is true in every possible circumstance where $t$ is true \cite{chierchia_meaning_2000}. This definition, as well as many other formal linguistic theories, is theoretically sound but practically too rigid to handle uncertainty. In practical NLP context, we define entailment to include cases where the truth value of hypothesis $h$ is highly plausible given text $t$, rather than absolutely certain \cite{dagan_recognizing_2009}. In other words, `text $t$ entails hypothesis $h$ if, typically, a human reading $t$ would infer that $h$ is most likely true' \cite{bar-haim_benchmarking_2014}. In contrast to a formal theoretical definition, this somewhat informal definition of entailment allows and requires common sense background knowledge.

The task of RTE started as a two-way classification of deciding whether hypothesis $h$ is entailed/supported by text $t$ or not \cite{bar-haim_second_2006, dagan_pascal_2005, giampiccolo_third_2007}. After the notion of `contradiction', i.e., `the negation of the hypothesis $h$ is entailed by the text $t$' is introduced \cite{de_marneffe_finding_2008}, the RTE task became a three-way classification task of predicting labels of a text pair out of `ENTAILMENT', `CONTRADICTION' and `UNKNOWN' \cite{giampiccolo_fourth_2008}. 

Driven by a yearly RTE challenge from 2005 to 2011, the NLP community developed some useful datasets for the task, specifically the RTE1 - RTE7 datasets. Despite being relatively small and imbalanced, these datasets enabled developing and evaluating of various approaches. While lexical-based and syntax-based approaches struggled to achieve good results \cite{bar-haim_benchmarking_2014, dagan_recognizing_2009}, machine learning approaches achieved reasonable performance, often combined with logical or probabilistic methods. 

One of the earlier models attempted to feed deep semantic features generated by first-order theorem prover and finite model builder into a machine learning classifie to make predictions \cite{bos_when_2006}. Surprisingly, the deep semantic features failed to outperform shallow semantic features. This is likely due to the models’ na\"ive and rigid representation of sentences, lack of background knowledge and flawed sample distribution of the dataset. 

Another intuitive approach is to first induce representations of text snippets into a hierarchical knowledge representation and then use a sound inferential mechanism to prove semantic entailment \cite{de_salvo_braz_inference_2006}. Despite its sound and tangible system design, this model only achieved an overall accuracy of $65.9\%$.

Furthermore, the NatLog system deals with the problem in three stages. It first conducts linguistic analysis, then aligns the dependency graph of the text $t$ and the hypothesis $h$, finally uses a decision tree classifier to perform entailment inference based on antonyms, polarity, graph structure and semantic relations \cite{chambers_learning_2007}. This NatLog system trades low recall (31.71 on RTE3 test set) for higher precision (68.06 on RTE3 test set).

To help address the low recall achieved by first-order rules, the class of pair feature spaces was introduced \cite{zanzotto_machine_2009}. It allowed the model to enrich the sentence-pair with `placeholders' and then generate first-order rewrite rules to relax the rigidness. This model achieved around $68\%$ overall accuracy on RTE3. 

Moreover, COGEX developed a system that first transforms the text into three-layered semantically-rich logic form representations, then generates a set of linguistic and world knowledge axioms, and searches for a proof of entailment \cite{tatu_cogex_2007}. This system achieved an overall accuracy of $72.25\%$. 

Overall, many inspiring hybrid models of logical inference and machine learning methods were developed for RTE challenges. Though they did not achieve perfect performance, we believe they have great potential once equipped with better text representations and more powerful neural models.

\paragraph{Natural Language Inference (NLI)}

More recently, NLI is proposed as `the problem of determining whether a natural language hypothesis $h$ can reasonably be inferred from a given premise $p$' \cite{bowman_large_2015, maccartney_extended_2009}. Noticeably, the definition of NLI is very similar to RTE and researchers tend to mention them together when addressing the problem. 

Despite that, NLI datasets have improvements over RTE datasets. Earlier RTE datasets, published before the notion of `CONTRADICTION' attracted enough attention, only have binary labelling of `ENTAILMENT'. In contrast, NLI datasets all include three-way labelling that includes `ENTAILMENT', `CONTRADICTION' and `UNKNOWN'. Furthermore, recent NLI datasets have larger size, more balanced label distribution and cover various domains. Table \ref{tab:CVdataset} presents NLI datasets that are potentially useful for claim verification.

With their large size and balanced design, NLI datasets have powered large neural network models, which has become the dominant approach. The common practice is to fine-tune a large pre-trained language model on the target NLI dataset, which may or may not be coupled with small task-specific techniques. Compared with traditional approaches, this approach improved text representations, achieved better generalisability and allowed more complex computing without relying on hand-crafted rules.

Current state-of-the-art models on NLI datasets are BERT \cite{devlin_bert_2019}, RoBERTa \cite{liu_roberta_2019}, MT-DNN \cite{liu_multi-task_2019}, ALBERT \cite{lan_albert_2020} and T5 \cite{raffel_exploring_2020}.

\section{Discussion and Challenges} \label{discussion}

In this section, we discuss current progress in each of the components of the automated fact-checking task, as well as highlight the main open challenges.

\subsection{Claim detection}
\paragraph{Conceptual Definition of Claim}
The definition of claim check-worthiness is brief \cite{allein2020checkworthiness}. Full Fact describe it as ``an assertion about the world that can be checked''. In contrast, \cite{konstantinovskiy_towards_2020} mentioned this definition is not enough to decide whether this claim is worthy for check or not. Similarly, \cite{factrank} declared that not every factual claim will be verified by fact checkers.

\paragraph{Narrow Domains}
Claims in the political domain are dominating the interest of journalists and researchers, as can be seen in existing datasets. As an example, \cite{wright2020claim} investigated the development of a claim check-worthiness detection method that would consistently perform over different domains, in this case rumours on Twitter, Wikipedia citations, and political speeches. However, the method showed important challenges in trying to perform well across domains. Recent research in claim detection has expanded to focus on health claims too, particularly owing to COVID-19.

\paragraph{Annotation Issues}
Labelling of sentences as claims or non-claims is generally done manually by non-experts (see Table \ref{tab:CDdataset}). An alternative to this is to derive labels from previously fact-checked claims collected from fact-checking websites. The main caveat of this approach is that fact-checking websites only list claims, rather than non-claims, which means that one needs to develop models that only leveraged instances of the positive class, i.e. positive unlabelled learning \cite{wright2020claim} .

\paragraph{Imbalanced Datasets}
The majority of datasets are imbalanced where not check-worthy claims outnumber check-worthy claims. While this is possibly due to the nature of the task, existing models can have a tendency to overfit due to this imbalance, which calls for more research to tackle the problem. For example, in the CheckThat! Lab 2020, \cite{accenture20} attempted to mitigate the problem of overfitting by retraining the model by resampling the larger number of positive instances that were augmenting data through translation between Arabic and English \cite{accenture20}.

\subsection{Claim Validation}
Despite the noticeable progress, current automated claim validation systems also face unique challenges and desire improvements over several key aspects: datasets quality, system integrity and model interpretability.

\paragraph{Datasets Quality}
State-of-the-art systems heavily rely on training large language models, which require large-scale, high-quality, labelled datasets that are expensive and may be unrealistic for specific domains. Despite the great contributions recent datasets have made, they tend to be imbalanced and somewhat synthetic, which are not ideal for model training. We believe future high-quality datasets will continue to help progress the field.

\paragraph{System Scalability and Integrity}
Proposed automated claim validation systems cover a range of relevant tasks. Though a few of them try to jointly handle rationale selection and claim verification \cite{hidey_team_2018, li_paragraph-level_2021}, most of them are pipeline systems that deal with subtasks separately. Improved scalability and integrity is desired.

Large pre-trained language models, the current dominant approach of various relevant tasks, requires lots of computing resources to train and inference. The scalability and accessibility of the proposed systems remain inferior.

Otherwise, increased system integrity is desired. Pipeline design has its inevitable disadvantage: downstream components can only make inferences on upstream results and errors accumulate throughout the pipeline. For instance, a claim verification component that takes the retrieved evidence and the detected claim as input will perform poorly with low-quality evidence or claims that are not checkable. Furthermore, the popular three-way label prediction approach is not the best approach for claim verification. Models struggle particularly to predict contradiction relation due to a lack of training data in this class, which accumulates errors across classes. For example, a model may predict a claim to be ``NO INFO'' while it should be ``CONTRADICT'', which makes it a false positive for the ``NO INFO'' class and a true negative for the ``CONTRADICT'' class. Preliminary research splitting the three-way classification into two binary classifications \cite{zeng2021qmulsds} is likely to help avoid such errors. Moreover, current approaches leave no space for aggregating evidence across sentences.

We believe a more compact overall system design is desired for automated claim validation such that it handles subtasks in a systematic way. We believe a promising direction is to train a model to learn all involved tasks in a multitask learning manner so that it may optimise for better overall performance.

\paragraph{Model Intepretability}

Neural networks are robust but struggle with interpretability and generalisability \cite{duan_machine_2020}, which is of particular importance for automated claim validation. Underwhelming model interpretability may induce an increased probability of models making the right prediction based on the wrong evidence. In contrast, symbolic systems that are unfortunately fragile and inflexible have strong interpretability and abstraction. Naturally, building a neural-symbolic system that integrates neural networks with symbolic logic becomes an interesting direction. In a nutshell, 
\textit{neural-symbolic systems = connectionist machine + logical abstractions 
\cite{besold_neural-symbolic_2017}}.

Researchers have proposed various architectures that incorporate first-order logic into neural networks. A recent study proposed a general framework capable of enhancing neural networks with declarative first-order logic \cite{hu_harnessing_2016}. Another study explored a symbolic intermediate representation for neural surface realisation \cite{elder_designing_2019} that is similar to first-order logic. Moreover, a recent attempt adapted module networks to model natural logic operations, which is enhanced with a memory component to model contextual information \cite{feng_exploring_2020}. Furthermore, RuleNN \cite{sen_learning_2020} is developed to tackle sentence classification where models are in the form of first-order logic, and achieved performance that is comparable to some neural models. 

Neural-symbolic methods have a fascinating potential of attaining interpretability from symbolic models and robustness from neural models makes. We believe that designing and implementing neural-symbolic methods for various tasks of automated fact-checking is promising and of particular interest to our society.

\section{Related Tasks}
\label{related}

There are some other popular tasks in natural language processing which are also related to the accuracy, verifiability and credibility of information, which we briefly discuss next as topics recommended for further reading:

\paragraph{Fake News Detection}
It is the task of determining whether a news article on the web is accurate or not \cite{shu2017fake}. Proposed classification approaches are typically centred on shallow features of the articles: n-grams, characters, stop words, part-of-speech tags, readability scores, term frequency, etc. Some more advanced approaches use additional metadata. However, these approaches are more likely to merely capture patterns of different article styles, rather than to sensibly distinguish reliable and unreliable articles \cite{hanselowski_machine-learning-based_2020}.

\paragraph{Rumour Detection}
It is the task of identifying unverified reports circulating on social media. Predictions are typically made on language subjectivity and metadata on social media \cite{zubiaga_detection_2018}. Despite the relevance of these features, the truth value of a claim does not directly depend on these features. 

\paragraph{Clickbait Detection}
Being considerably different from automated fact-checking, clickbait detection does not require external evidence. Approaches with relatively shallow linguistic features \cite{chakraborty_stop_2016, chen_misleading_2015, potthast_clickbait_2016} have yielded reasonable performance. 

\paragraph{Commonsense Reasoning}
To perform commonsense reasoning \cite{storks2019commonsense}, the model needs to be able to do reasoning beyond the explicit information given in sentence pairs, which is highly valued in automated fact-checking \cite{thorne_automated_2018}. As a new frontier of artificial intelligence, novel studies have investigated learned knowledge in pre-trained language models, commonsense integration from external knowledge bases, symbolic knowledge incorporation, etc. However, these tasks are currently under investigation and the field calls for major breakthroughs. For more information, we refer to a recent survey \cite{storks_commonsense_2019} and a tutorial \cite{sap_commonsense_2020}.

\section{Conclusions} \label{conclusions}

In this paper, we present a survey on automated fact-checking with special focus on claim detection and claim validation. Substantial progress has been made by applying large pre-trained language models through designed pipelines, but numerous open challenges still need further research. Claim Detection faces challenges from conceptual definition, narrow domains, annotation issues and imbalanced datasets. In addition, improvements over datasets quality, system integrity and model intepretability are desired for claim validation.

\section*{Acknowledgments}

This work was supported by the Engineering and Physical Sciences Research Council (grant EP/V048597/1). Xia Zeng is funded by China Scholarship Council (CSC). Amani S. Abumansour holds a scholarship from Taif University, Saudi Arabia.

\bibliographystyle{apalike}

\bibliography{references}

\end{document}